# Kriging Interpolation Filter to Reduce High Density Salt and Pepper Noise

Firas Ajil Jassim

Management Information System Department
Irbid National University
Irbid, Jordan

*Abstract*—Image denoising is a critical issue in the field of digital image processing. This paper proposes a novel Salt & Pepper noise suppression by developing a Kriging Interpolation Filter (KIF) for image denoising. Gray-level images degraded with Salt & Pepper noise have been considered. A sequential search for noise detection was made using k×k window size to determine non-noisy pixels only. The non-noisy pixels are passed into Kriging interpolation method to predict their absent neighbor pixels that were noisy pixels at the first phase. The utilization of Kriging interpolation filter proves that it is very impressive to suppress high noise density. It has been found that Kriging Interpolation filter achieves noise reduction without loss of edges and detailed information. Comparisons with existing algorithms are done using quality metrics like PSNR and MSE to assess the proposed filter.

## I. INTRODUCTION

Image noise may be defined as any corrosion in the image signal, caused by external disturbance. Thus, one of the most important areas of image restoration is that cleaning an image spoiled by noise. Digital images are often corrupted by impulse noise also known as Salt and Pepper noise due to transmission errors [14]. The goal of noise reduction is to detect noisy pixels and substitute an efficient (predicted) value for each and this is however the truly definition of filters. The way in which the pixel is estimated as noisy or not noisy depends on how estimate is calculated [9]. The most commonly used filters are the Standard Median Filter (SMF), Adaptive Median Filter (AMF) [5], Decision Based Algorithm (DBA) [8], Progressive Switching Median Filter (PSMF) [22], and Detail preserving filter (DPF) [11]. The filtering algorithm varies from one algorithm to another by the approximation accuracy for the noisy pixel from its surrounding pixels [3]. Among these the Median Filter (MF) is used widely because of its effective noise suppression capability [19]. Practically, one of the main disadvantages of the Median filter is that it modifies both noisy and non-noisy pixels thus removing some fine details of the image. Hence, to overcome this disadvantage, the Adaptive Median Filter (AMF) was proposed [5]. AMF perform well at low noise densities but at high noise densities the window size has to be increased which may leads to blurring the image [9]. However, when the noise level is over 50%, some details and edges of the original image are contaminated by the filter [17]. Therefore it is only suitable for low level noise density. At high noise density it shows the blurring for the larger window sizes and not able to suppress the noise completely [16].

Nowadays, two types of filtering techniques could be exploited in image denoising which are linear and non-linear filters. For high noise density, the output images are blurred and edges are not preserved accurately by the linear filters. Alternatively, the non-linear filters have been used to provide better filtering performance in terms of impulse noise removal and preservation of other details of the images [16].

The outline of the paper is as follows: The definition of impulse noise (Salt & Pepper) was reviewed in Section II. Kriging interpolation method was introduced in Section III. The proposed Kriging filter was presented in Section IV. Experimental results and conclusions are presented in Sections V and VI, respectively.

## II. IMAGE NOISE PRELIMINIRAIES

Image denoising is the process of finding unusual values in digital image, which may be the result of errors made by external effects in image capturing process. Many text books in image processing include chapters about image noise and enhancement [7][14][21]. Noise represents unwanted information which spoils image quality. Also, noise may be defined as pixels that are different from their neighbors and they are markedly noticeable by human eye [21]. According to [14], the noise can be written as:

$$f(x, y) = g(x, y) + \eta(x, y) \qquad (1)$$

where f(x,y) is the original image and g(x,y) is the output image plus $\eta$(x,y) which is the noise model. Now, a mathematical transformation T could be introduced as follows:

$$g(x, y) = T\big[f(x, y)\big] \qquad (2)$$

T is an operator on f where the fundamental operation of the operator T that it works as a transformation to transform the original image f(x,y) from the original state (noisy state) into g(x,y) which is the output state (non-noisy state) [15].

### A. Salt & Pepper Noise

The Salt & Pepper noise is generally caused by defect of camera's sensor, by software failure or by hardware failure in image capturing or transmission. Due to this situation, Salt & Pepper noise model, only a proportion of all the image pixels are corrupted whereas other pixels are non-noisy [22]. Since, in this paper, gray-level images degraded with Salt & Pepper noise have been considered, the noise value may be either minimum (0) or maximum (255) of the gray scale of the image. For an 8 bit/pixel image, the typical intensity value for pepper noise is close to 0 and for salt noise is close to 255. Furthermore, the unaffected pixels remain unchanged.

$$\eta(x, y) = \begin{cases} 0, & \text{Pepper noise} \\ 255, & \text{Salt noise} \end{cases} \qquad (3)$$

### III. KRIGING INTERPOLATION

Kriging is a statistical technique permitting to estimate unknown values at specific points in space by using data values from known locations. Kriging provides exact interpolation, i.e., predicted output values at inputs already observed equal the observed output values [13]. Kriging produces optimal results compared with other interpolation techniques [10]. Furthermore, Kriging is attractive because it can ensure that the prediction has exactly the same value as the observed one [20]. Actually, these predictions are weighted linear combinations of the observed values. Kriging assumes that the closer the input data, the more positively correlated the prediction errors [20]. Obviously, the pixels within the k×k block size are highly correlated [14], therefore, the application of Kriging within the same block will produce more positively correlated predictions. Kriging confer weights for each point according to its distance from the unobserved value. Actually, these predictions treated as weighted linear combinations of the known values. The weights should provide a Best Linear Unbiased Estimator (BLUE) or Best Linear Unbiased Predictor (BLUP) of the output value for a given input [12]. The fundamental advantage of kriging over traditional interpolation methods is that it uses the spatial correlation structure of the data set being interpolated in order

to calculate the unobserved estimate [13]. The general form of Kriging method is as follows:

$$\hat{Z}^* = \sum_{i=1}^{N} \lambda_i Z_i \qquad (4)$$

where

$$\sum_{i=1}^{N} \lambda_i = 1$$

The kriging estimate is obtained by choosing $\lambda_i$'s that minimize variance of the estimator under the unbiasedness constraint:

$$\sigma^2 = E[(Z - \hat{Z}^*)^2]$$

There are several Kriging types, differ in their treatments of the weighted components $\lambda_i$'s. Here, in the proposed technique, ordinary kriging will be used due to the fact that it is the most common kriging type and it is considered to be best because it minimizes the variance of the estimation error [20].

In Kriging, a significant role is played by the variogram: A diagram of the variance of the difference between the measurements at two input locations. The variogram describes the variance of the difference of samples within the data set and is calculated by the following equation:

$$2\gamma(h) = \frac{1}{n} \sum_{i=1}^{n} \big[z(x_i) - z(x_i + h)\big]^2 \qquad (5)$$

where z(x) is the value of the data at point x and z(x+h) is the value at a point with a lag distance h from x. The semi-variogram is one-half the value of the variogram. This article focuses of the Kriging as a scheme embedded in image denoising and there is no need to explain the behavior and the analysis of the variogram and semi-variogram. A detailed discussion about variogram analysis and estimation could be found through recommended readings [4][2][4][13].

### IV. PROPOSED KRIGING FILTER

First of all, many authors [6][10][18] turned to implement Kriging in image restoration but they all run after Gaussian noise and no steady work could be found about the use of Kriging interpolation to suppress Salt & Pepper noise for image restoration. According to [6], Kriging filter for Gaussian noise is clearly superior to other non-statistical procedures. Thus, from this point the proposed Kriging filter arise to deal with the problem of suppressing Salt & Pepper noise in gray scale images. Therefore, the noise considered by the proposed algorithm is only Salt & Pepper noise.

At the beginning, the problem of detecting noisy pixels exactly is very important to preserve edges and original image information in the best way possible. Obviously, Salt & Pepper noise is either 0 or 255, therefore; a sequential search could be implemented in the k×k window size. The pattern of the sequential search is to find non-noisy pixels, i.e. pixels that are neither 0 nor 255. Here, the proposed filter will conflict with all the earlier and recent researches in this field since all the work carried out was by searching the noisy pixels and substituting a suitable value for it. The idea behind searching non-noisy pixels is to construct three vectors $X \in R^N$, $Y \in R^N$, and $Z \in R^N$ where N is the number of non-noisy pixel in the k×k window size. The vectors X and Y represent the location of the non-noisy pixel along the horizontal and vertical direction whereas Z vector used to store the gray level pixel values that lie between 0 used to 255, exclusive, i.e, 1 to 254. An illustrative example would be helpful to understand the above idea. Hence, an arbitrary 3×3 noisy window size has been considered in figure (1).

|   | 1 | 2 | 3 |
|---|---|---|---|
| 1 | 0 | 88 | 85 |
| 2 | 88 | 255 | 0 |
| 3 | 255 | 88 | 86 |

Figure 1.   Noisy 3×3window size

However, there are five non-noisy pixels and their locations are (1,2), (1,3), (2,1), (3,2), and (3,3). Hence, the X vector will contains the location of non-noisy pixels in horizontal direction as X={1,1,2,3,3} while Y={2,3,1,2,3} using the same procedure along the vertical direction. The values of the original window for non-noisy pixels will be stored in Z as {88, 85, 88, 88, 86}. Subsequently, Kriging interpolation could be applied to the X, Y, and Z vectors using a mesh grid $X_0$=1,2,3 and $Y_0$=1,2,3 to predict the values in the (1,1), (2,2), (2,3), and (3,1) locations which are the locations of the noisy pixels, figure (2).

|   | 1 | 2 | 3 |
|---|---|---|---|
| 1 | $Z^*_1$ | 88 | 85 |
| 2 | 88 | $Z^*_2$ | $Z^*_3$ |
| 3 | $Z^*_4$ | 88 | 86 |

Figure 2.   Identifying noisy pixels

According to (4), the value of $Z^*_1$ could be computed as:

$$Z^*_1 = \lambda_1(88) + \lambda_2(85) + \lambda_3(88) + \lambda_4(88) + \lambda_5(86)$$

As a result, after applying Kriging interpolation technique, the values of the weights are $\lambda_1$=0.2030, $\lambda_2$=0.1976, $\lambda_3$=0.2041, $\lambda_4$=0.1976, and $\lambda_5$=0.1976 and it is truly that $\sum \lambda$'s=1. Then, the value of the unobserved value is $Z^*_1$=87.0119≈87. A similar technique could be used to predict the values of $Z^*_2$, $Z^*_3$, and $Z^*_4$. Hence, the 3×3 predicted block that is noise free could be represented as:

|   | 1 | 2 | 3 |
|---|---|---|---|
| 1 | 87 | 88 | 85 |
| 2 | 88 | 87 | 87 |
| 3 | 87 | 88 | 86 |

Figure 3.   Predicting noisy pixels using Kriging

Similarly, the same previous procedure may be applied using k×k window size to the whole image to obtain the exact result as figure (3). It must be mentioned that it is just a coincidence that the values of $Z^*_1$, $Z^*_2$, $Z^*_3$, and $Z^*_4$ are equal. This is because the small 3×3 window size and it would be different for large window size and another arbitrary k×k block.

## V.   EXPERIMENTAL RESULTS

In order to demonstrate the performance of the proposed filter, an 8-bit gray scale 512×512 Lena image contaminated by Salt & Pepper noise with noise occurrence of 10% to 90% is considered. The performance of the Kriging interpolation filter (KIF) was tested against Standard Median Filter (SMF), Adaptive Median Filter (AMF), Progressive Switching Median Filter (PSMF), Detail preserving filter (DPF), Decision Based Algorithm (DBA). As a result, the proposed filter was found to perform quite well on image corrupted with high Salt & Pepper noise up to the level of 90%. The Peak Signal to Noise Ratio (PSNR) and Mean Square Error (MSE), are used to measure the objective dissimilarities between the filtered images and the original image.

$$PSNR = 10 \, Log_{10} \frac{255^2}{MSE}$$

$$MSE = \frac{1}{nm} \sum_{i=1}^{n} \sum_{j=1}^{m} (f_{ij} - g_{ij})^2$$

According to [1], it is desirable to work with large windows, since they obviously contain more information than small ones. Therefore, to figure out the suitable window size for Kriging filter, a comparison of different window sizes 4×4, 8×8, and 16×16 have been established. These results show that the PSNR value for 4×4 window size was low. Moreover, according to figure (4), there is a very small difference between 8×8 and 16×16 window sizes. Therefore, an 8×8 window size was chosen in this paper because it is fulfill the same result as 16×16 window size.

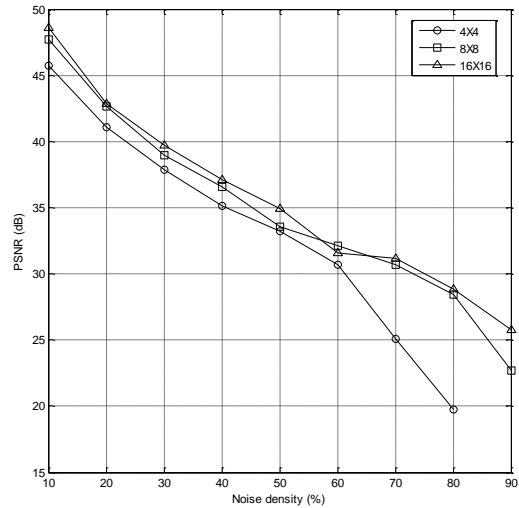

Figure 4.   Multiple window Sizes for Kriging Filter

Moreover, to support the decision of choosing 8×8 window size, a comparison between three window sizes of Kriging versus computation time has been made. Clearly, for high noise densities (50% ~ 90%), 8×8 window size gives better results that is less computation time than 4×4 and 16×16 windows. Since this article concerning high noise densities, then 8×8 window size will be considered as fair window size for this paper.

According to tables I & II, the values of PSNR and MSE were demonstrated for the proposed Kriging filter. Clearly, the proposed technique outperforms the other filters, in terms of PSNR and MSE. Moreover, figure (6) and (7), mimic the graphical differences between KIF and the other filters in terms of PSNR and MSE, respectively. Moreover, ocular restoration of Lena image concerning KIF using different noise densities varying (10% ~ 90%) are presented in figure (8). For more clarity, a magnified ocular dissimilarities, same as figure (8), are presented in figure (9). In figure (10), visual dissimilarities between KIF and other noise removal filters for the highest image noise which is 90% are presented. The restorations after contamination with 90% noise are clearly sound that KIF is dominating the other filters in its noticeable result.

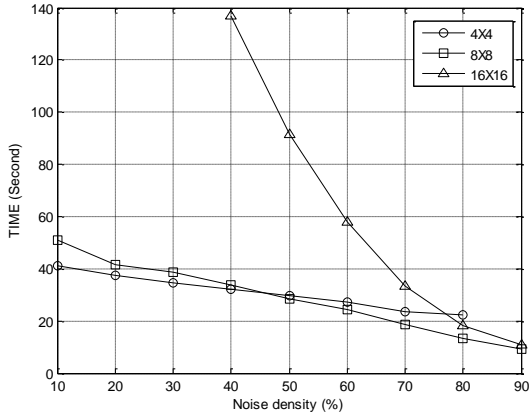

Figure 5.   Computation time for multiple window Sizes for Kriging Filter

TABLE I.        PSNR (dB) FOR DIFFERENT DENOISING METHODS FOR LENA (512×512) IMAGE

| Noise Ratio (%) | Kriging | SMF | AMF | PSM | DPF | DBA |
|---|---|---|---|---|---|---|
| 10 | 47.68 | 34.92 | 39.38 | 38.85 | 33.81 | 46.66 |
| 20 | 42.63 | 30.3 | 36.93 | 33.41 | 27.53 | 41.72 |
| 30 | 38.97 | 23.99 | 34.68 | 29.4 | 23.18 | 38.04 |
| 40 | 36.57 | 19.02 | 32.27 | 25.45 | 19.76 | 36.00 |
| 50 | 33.58 | 15.93 | 27.38 | 25.39 | 16.8 | 32.38 |
| 60 | 32.09 | 12.36 | 21.66 | 21.27 | 14.51 | 29.82 |
| 70 | 30.67 | 10.08 | 16.6 | 9.94 | 12.5 | 24.30 |
| 80 | 28.40 | 8.15 | 12.79 | 8.1 | 10.79 | 19.47 |
| 90 | 22.69 | 6.6 | 9.86 | 6.68 | 9.21 | 12.27 |

TABLE II.        MSE FOR DIFFERENT DENOISING METHODS FOR LENA (512×512) IMAGE

| Noise Ratio (%) | Kriging | SMF | AMF | PSM | DPF | DBA |
|---|---|---|---|---|---|---|
| 10 | 1.98 | 20.90 | 7.40 | 8.40 | 27.00 | 4.49 |
| 20 | 4.56 | 60.60 | 13.10 | 29.60 | 114.60 | 11.94 |
| 30 | 8.67 | 259.30 | 22.10 | 74.50 | 312.10 | 22.88 |
| 40 | 13.45 | 814.20 | 38.50 | 185.20 | 686.90 | 38.67 |
| 50 | 25.09 | 1877.90 | 118.70 | 187.50 | 1355.80 | 63.83 |
| 60 | 35.15 | 3776.30 | 443.20 | 484.20 | 1197.80 | 103.18 |
| 70 | 49.79 | 637.90 | 1421.20 | 600.00 | 3651.20 | 175.11 |
| 80 | 83.70 | 9945.80 | 3413.70 | 1000.60 | 5408.90 | 314.50 |
| 90 | 317.81 | 14179.00 | 6708.80 | 1396.00 | 7798.40 | 772.21 |

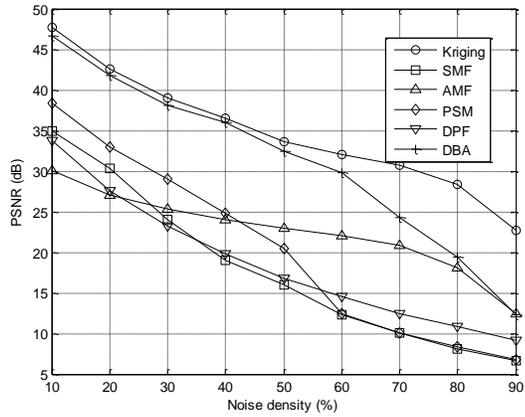

Figure 6.   PSNR (dB) for different denoising methods

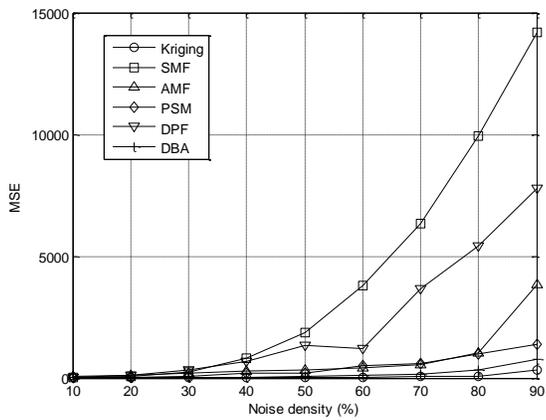

Figure 7.   MSE for different denoising methods

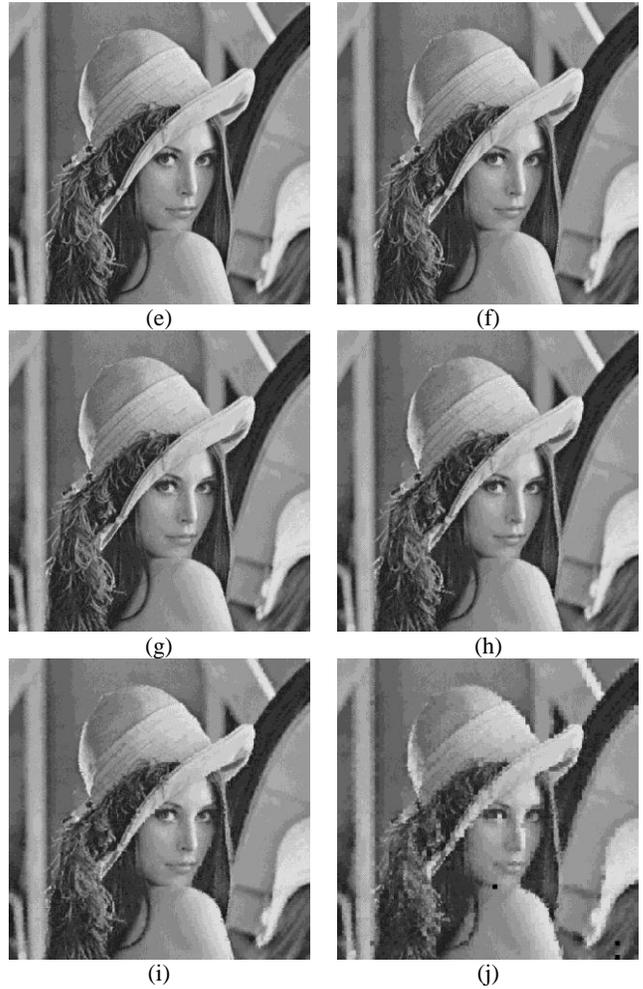

Figure 8.   Restoration with Kriging filter (a) Original (b) After 10% (C) After 20% (d) After 30% (e) After 40% (f) After 50% (g) After 60% (h) After 70% (i) After 80% (j) After 90%

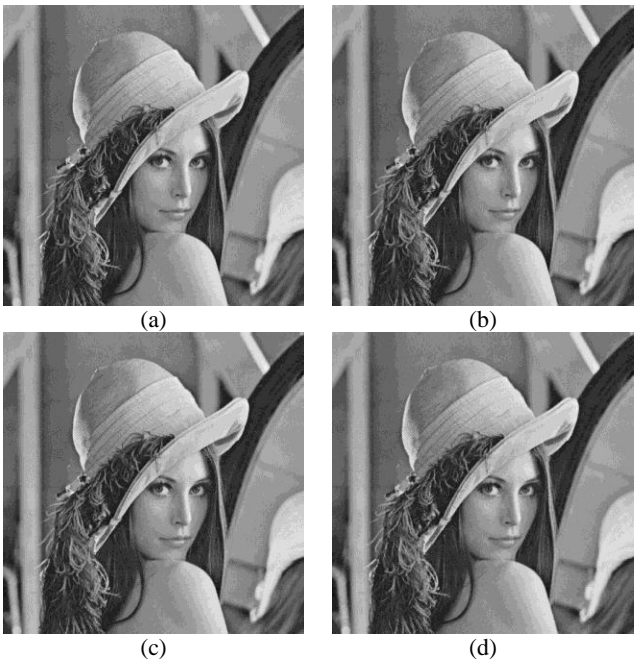

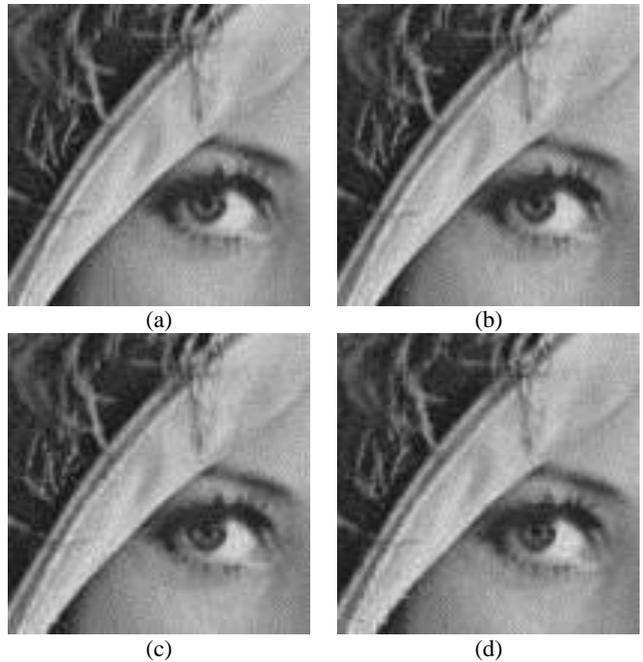

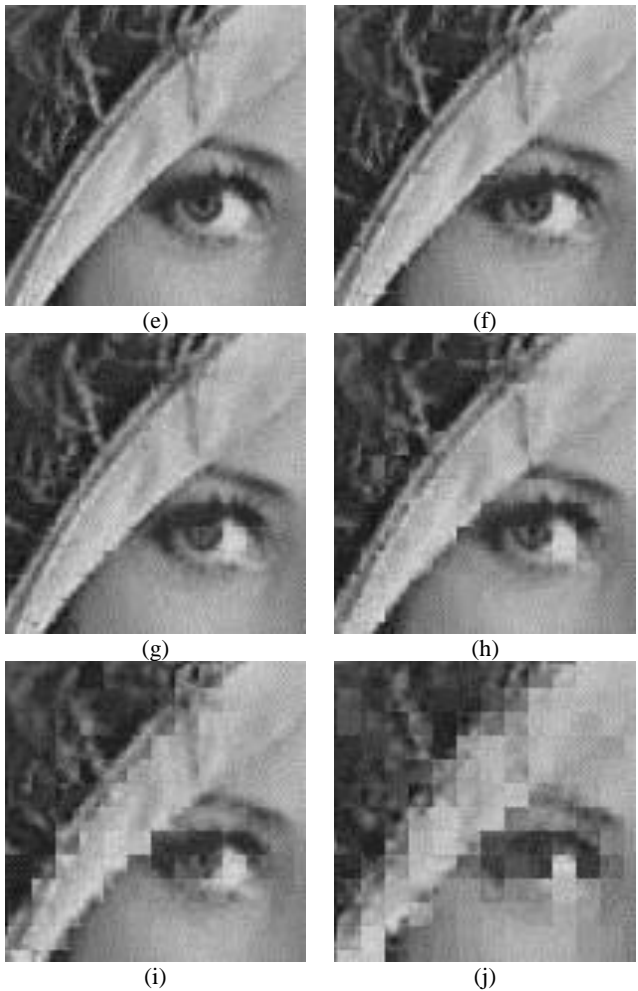

Figure 9. A magnified restoration with Kriging filter (a) Original (b) after 10% (C) After 20% (d) After 30% (e) After 40% (f) After 50% (g) After 60% (h) After 70% (i) After 80% (j) After 90%

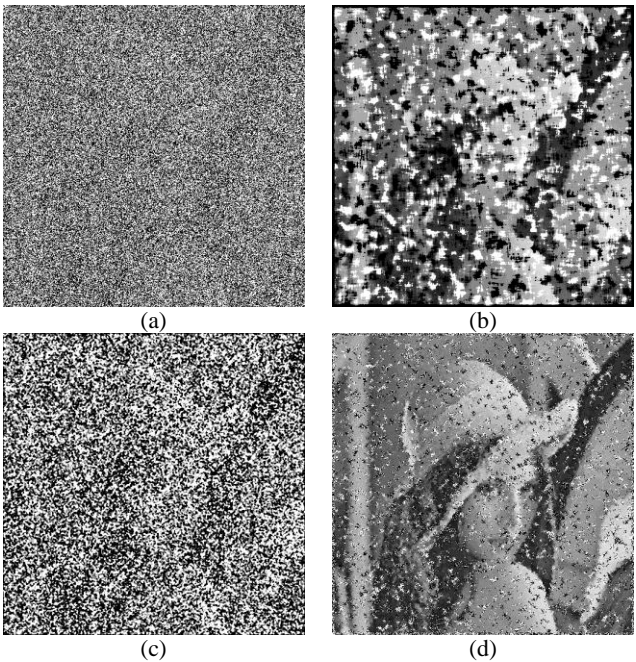

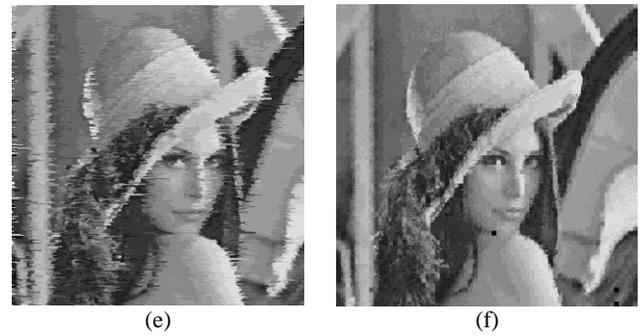

Figure 10. (a) 90% noise (b) restoration with MF (c) restoration with PSMF (d) restoration with AMF (e) restoration with DBA (f) restoration with proposed Kriging filter

## VI. CONCLUSIONS

In this paper, a novel statistical filter for image denoising has been established using Kriging interpolation method. The proposed Kriging filter gives better performance in comparison with other available filters in the field. Experimental results reveal that the proposed Kriging filter significantly outperforms other techniques by having higher PSNR and lesser MSE. Furthermore, the visual results of the proposed filter across a wide range of noise densities, varying from 10% to 90%, show that Kriging filter provides maximum suppression of impulse noise while preserving edges and fine details.

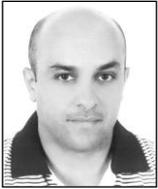

**Firas A. Jassim** received the BS degree in mathematics and computer applications from Al-Nahrain University, Baghdad, Iraq in 1997, and the MS degree in mathematics and computer applications from Al-Nahrain University, Baghdad, Iraq in 1999 and the PhD degree in computer information systems from the university of banking and financial sciences, Amman, Jordan in 2012. Now, he is working as an assistant professor with Management Information System Department at Irbid National University, Irbid, Jordan. His research interests are Image processing, image compression, image enhancement, image interpolation and simulation.